\documentclass[conference]{IEEEtran}
\usepackage[pdftex]{graphicx}
\usepackage{acronym}
\usepackage{multirow}
\usepackage{amsmath, amsthm, amssymb, amsfonts, mathtools}

\usepackage{xcolor}

\begin{document}
%
\title{On-chip Few-shot Learning with Surrogate Gradient Descent on a Neuromorphic Processor}

\providecommand{\keywords}[1]
{
  \small	
  \textbf{\textit{Keywords---}} #1
}

\acrodef{AC}[AC]{Arrenhius \& Current}
\acrodef{AER}[AER]{Address Event Representation}
\acrodef{AEX}[AEX]{AER EXtension board}
\acrodef{AMDA}[AMDA]{``AER Motherboard with D/A converters''}
\acrodef{API}[API]{Application Programming Interface}
\acrodef{BM}[BM]{Boltzmann Machine}
\acrodef{CAVIAR}[CAVIAR]{Convolution AER Vision Architecture for Real-Time}
\acrodef{CCN}[CCN]{Cooperative and Competitive Network}
\acrodef{CD}[CD]{Contrastive Divergence}
\acrodef{CMOS}[CMOS]{Complementary Metal--Oxide--Semiconductor}
\acrodef{COTS}[COTS]{Commercial Off-The-Shelf}
\acrodef{CPU}[CPU]{Central Processing Unit}
\acrodef{CV}[CV]{Coefficient of Variation}
\acrodef{CV}[CV]{Coefficient of Variation}
\acrodef{DAC}[DAC]{Digital--to--Analog}
\acrodef{DBN}[DBN]{Deep Belief Network}
\acrodef{DFA}[DFA]{Deterministic Finite Automaton}
\acrodef{DFA}[DFA]{Deterministic Finite Automaton}
\acrodef{divmod3}[DIVMOD3]{divisibility of a number by 3}
\acrodef{DPE}[DPE]{Dynamic Parameter Estimation}
\acrodef{DPI}[DPI]{Differential-Pair Integrator}
\acrodef{DSP}[DSP]{Digital Signal Processor}
\acrodef{DVS}[DVS]{Dynamic Vision Sensor}
\acrodef{EDVAC}[EDVAC]{Electronic Discrete Variable Automatic Computer}
\acrodef{EIF}[EI\&F]{Exponential Integrate \& Fire}
\acrodef{EIN}[EIN]{Excitatory--Inhibitory Network}
\acrodef{EPSC}[EPSC]{Excitatory Post-Synaptic Current}
\acrodef{eRBP}[eRBP]{event-driven Random Back-Propagation}
\acrodef{EPSP}[EPSP]{Excitatory Post--Synaptic Potential}
\acrodef{FPGA}[FPGA]{Field Programmable Gate Array}
\acrodef{FSM}[FSM]{Finite State Machine}
\acrodef{GPU}[GPU]{Graphical Processing Unit}
\acrodef{HAL}[HAL]{Hardware Abstraction Layer}
\acrodef{HH}[H\&H]{Hodgkin \& Huxley}
\acrodef{HMM}[HMM]{Hidden Markov Model}
\acrodef{HW}[HW]{Hardware}
\acrodef{hWTA}[hWTA]{Hard Winner--Take--All}
\acrodef{IF2DWTA}[IF2DWTA]{Integrate \& Fire 2--Dimensional WTA}
\acrodef{IF}[I\&F]{Integrate \& Fire}
\acrodef{IFSLWTA}[IFSLWTA]{Integrate \& Fire Stop Learning WTA}
\acrodef{INCF}[INCF]{International Neuroinformatics Coordinating Facility}
\acrodef{INI}[INI]{Institute of Neuroinformatics}
\acrodef{IO}[IO]{Input-Output}
\acrodef{IPSC}[IPSC]{Inhibitory Post-Synaptic Current}
\acrodef{ISI}[ISI]{Inter--Spike Interval}
\acrodef{JFLAP}[JFLAP]{Java - Formal Languages and Automata Package}
\acrodef{LIF}[LI\&F]{Leaky Integrate \& Fire}
\acrodef{LSM}[LSM]{Liquid State Machine}
\acrodef{LTD}[LTD]{Long-Term Depression}
\acrodef{LTI}[LTI]{Linear Time-Invariant}
\acrodef{LTP}[LTP]{Long-Term Potentiation}
\acrodef{LTU}[LTU]{Linear Threshold Unit}
\acrodef{MCMC}{Markov Chain Monte Carlo}
\acrodef{MSE}[MSE]{Mean-Squared Error}
\acrodef{NHML}[NHML]{Neuromorphic Hardware Mark-up Language}
\acrodef{NMDA}[NMDA]{NMDA}
\acrodef{NME}[NE]{Neuromorphic Engineering}
\acrodef{PCB}[PCB]{Printed Circuit Board}
\acrodef{PRC}[PRC]{Phase Response Curve}
\acrodef{PSC}[PSC]{Post-Synaptic Current}
\acrodef{PSP}[PSP]{Post--Synaptic Potential}
\acrodef{RI}[KL]{Kullback-Leibler}
\acrodef{RNN}[RNN]{Recurrent Neural Network}
\acrodef{RRAM}[RRAM]{Resistive Random-Access Memory}
\acrodef{RBM}[RBM]{Restricted Boltzmann Machine}
\acrodef{ROC}[ROC]{Receiver Operator Characteristic}
\acrodef{SAC}[SAC]{Selective Attention Chip}
\acrodef{SCD}[SCD]{Spike-Based Contrastive Divergence}
\acrodef{SCX}[SCX]{Silicon CorteX}
\acrodef{SSM}[SSM]{Synaptic Sampling Machines}
\acrodef{SNN}[SNN]{Spiking Neural Network}
\acrodef{STDP}[STDP]{Spike Time Dependent Plasticity}
\acrodef{SW}[SW]{Software}
\acrodef{sWTA}[SWTA]{Soft Winner--Take--All}
\acrodef{VHDL}[VHDL]{VHSIC Hardware Description Language}
\acrodef{VLSI}[VLSI]{Very  Large  Scale  Integration}
\acrodef{WTA}[WTA]{Winner--Take--All}
\acrodef{XML}[XML]{eXtensible Mark-up Language}

\author{\IEEEauthorblockN{Kenneth Stewart}
\IEEEauthorblockA{Computer Science Department\\
UC, Irvine\\
Irvine, CA 92697-2625 USA\\
kennetms@uci.edu}
\and
\IEEEauthorblockN{Garrick Orchard}
\IEEEauthorblockA{Intel Labs\\Intel Corporation\\
Santa Clara, California\\
garrick.orchard@intel.com}
\and
\IEEEauthorblockN{Sumit Bam Shrestha}
\IEEEauthorblockA{Tamasek Laboratories @ NUS \\
National University of Singapore\\
Singapore\\
tslsbs@nus.edu.sg}
\and
\IEEEauthorblockN{Emre Neftci}
\IEEEauthorblockA{Cognitive Sciences Department\\and\\Computer Science Department\\
UC, Irvine\\
Irvine, CA 92697-2625 USA\\
eneftci@uci.edu}
}

\maketitle

\begin{abstract}
 Recent work suggests that synaptic plasticity dynamics in biological models of neurons and neuromorphic hardware are compatible with gradient-based learning \cite{Neftci_etal19_surrgrad}.
 Gradient-based learning requires iterating several times over a dataset, which is both time-consuming and constrains the training samples to be independently and identically distributed.
 This is incompatible with learning systems that do not have boundaries between training and inference, such as in neuromorphic hardware.
 One approach to overcome these constraints is transfer learning, where a portion of the network is pre-trained and mapped into hardware and the remaining portion is trained online. 
 Transfer learning has the advantage that pre-training can be accelerated offline if the task domain is known, and few samples of each class are sufficient for learning the target task at reasonable accuracies.
 Here, we demonstrate on-line surrogate gradient few-shot learning on Intel's Loihi neuromorphic research processor using features pre-trained with spike-based gradient backpropagation-through-time. 
 Our experimental results show that the Loihi chip can learn gestures online using a small number of shots and achieve results that are comparable to the models simulated on a conventional processor.

\end{abstract}

\hspace{8pt}



%
\IEEEpeerreviewmaketitle

\section{Introduction}
Understanding how the plasticity dynamics in multilayer biological neural networks are organized for efficient data-driven learning is a long-standing question in computational neuroscience \cite{Gerstner_etal14_neurdyna}, and elucidating it can open the door to scalable brain-inspired computers for continual ``life-long'' learning.
Machine learning and artificial neural networks provide important hints for answering this question.
Recent work has demonstrated that many ingredients of deep learning are compatible with biological neural networks and neuromorphic hardware \cite{Neftci18_datapowe,Neftci_etal19_surrgrad}. 
This compatibility paves the road towards neuromorphic learning machines with performances similar to deep learning, while being able to learn online, with input streams, and using a fraction of the energy.

However, neural network training is especially time-consuming in neuromorphic hardware since they generally operate in real-time and on a single stream of data (i.e. in deep learning terms, using a batchsize of 1).
Furthermore, the training samples must be independently and identically distributed (iid).
The failure to meet the iid constraint often results in catastrophic forgetting \cite{McClelland_etal95_whyther}. 
Methods such as transfer learning \cite{Yosinski_etal14_howtran} and meta-learning \cite{Finn_etal17_modemeta} can mitigate these two problems by pre-training a network in an iid fashion on a known task domain, followed by a small number of ``shots'' to learn the task-specific output.
Catastrophic forgetting is mitigated, because fewer data samples are needed to achieve a target accuracy compared to training from scratch, which translates into fewer synaptic weight updates.

Here, we leverage recent advances in surrogate gradient learning \cite{Neftci_etal19_surrgrad} and on-chip plasticity in neuromorphic hardware to demonstrate few-shot learning in an event-based gesture recognition task in Intel's neuromorphic research processor, Loihi.
A spiking convolutional neural network is trained offline using backpropagation through time \cite{Shrestha_Orchard18_slayspik} using an accurate behavioral model of the Loihi neuron.
The classification layer is then learned on-line using a variation of a three-factor rule for surrogate gradient learning \cite{Zenke_Ganguli17_supesupe} adapted to overcome limitations in neuromorphic hardware such as locality and bit precision.
We test the network learning performance on a gesture recognition task in a few-shot learning framework, and the performance is compared to that of an equivalent software model trained and executed on a conventional computer.
Our results demonstrate that the Loihi hardware achieves similar results to that of the software simulations, and provided the building blocks for multilayer learning and transfer learning in neuromorphic hardware.

\section{Surrogate Gradient Descent with the Loihi Neuromorphic Research Chip}

\subsection{Neuron and Plasticity Model}
The dynamics of the Loihi neuron follow the current-based leaky integrate-and-fire model.
The neuron uses two internal state variables: 
The synaptic response current $u_i[n]$ and the membrane potential $v_i[n]$. The current is computed as the sum of filtered input spike trains added to a constant bias current; 
\begin{equation}
    \begin{split}
        u_i[n] &= \sum_j w_{ij}Q_j[n]\\ 
        Q_j[n+1] &= \alpha_Q Q_j[n] + \frac{1}{\tau_u}S_j[n].
    \end{split}
\end{equation}
The variable $S_j[n]$ is the input spike train, modeled as a sum of boxcar functions of width and height equal to 1, with $n$ the discrete time step. 
$Q_j[n]$ here is a first-order low-pass filter that models the unweighted post-synaptic current. 
Upon an input spike $S_j$, $u_i$ undergoes a jump of height $w_{ij}$, then decays exponentially with time constant $\tau_u$, where $\alpha_Q=\exp(-\frac{1}{\tau_u})$.
The connection weights $w_{ij}$ are the trainable parameters of the neuron.
This synapse model is commonly used in computational neuroscience modeling \cite{Destexhe_etal98_methneur,Gerstner_etal14_neurdyna}. 
In the equations above, inputs neurons are confounded with network neurons for simplicity of notation. 
Although recurrent connections can be learned using the same dynamics as described in this work, our experiments focus on training feed-forward weights only.
 
The membrane potential follows similar dynamics and is the integration of the synaptic current with an exponential decay. 
A neuron emits a spike when its membrane potential exceeds the firing threshold $v_{th}$.
\begin{equation}\label{eq:loihi_dynamics}
    \begin{split}
        S_i[n]   &= \Theta(v_i[n]-v_{th}),\\
        v_i[n]   &= \sum_j w_{ij}P_j[n] + R_i[n],\\
    \end{split}
\end{equation}
where, 
\begin{equation}\label{eq:loihi_dynamics2}
    \begin{split}
        P_j[n+1] &= \alpha_P P_j[n] + \frac{1}{\tau_v}Q_j[n],\\
        R_i[n+1] &= \alpha_R R_i[n] - S_i v_{th}.
    \end{split}
\end{equation}
The term $\alpha_P = \exp(-\frac{1}{\tau_v})$ captures the leak of state $v$. 
Following an input spike $S_j$, $P$ follows an unweighted ``alpha'' post-synaptic potential with fast rise and slow decay. 
The term $R_i[n]$ captures the reset of the neuron after firing.  
The neuron model above is a discrete-time version of the Spike Response Model (SRM$_0$) \cite{Gerstner_etal14_neurdyna} and is fully compatible with the Loihi neuron model.

The Loihi research processor offers a variety of local information to a programmable synaptic learning process including pre and post synaptic traces and reward traces corresponding to special reward spikes \cite{Davies_etal18_loihneur}. 
Synaptic weights can be updated via a learning rule expressed as a finite-difference equation with respect to a synaptic state variable that follows a sum-of-products form as follows \cite{Davies_etal18_loihneur}:
\begin{equation}\label{eq:loihi_rule}
    w_{ij}[n+1] = w_{ij}[n]+\sum_k C_{k}\prod_l F_{kl},
\end{equation}
where $w_{ij}$ is the synaptic weight variable defined for the destination-source neuron pair being updated; $C_k$ is a scaling constant; and $F_{kl}$ may represent a synaptic state variable, a pre-synaptic trace, or a post-synaptic trace defined for the neuron pair $i$ and $j$. 
The weights are stochastically rounded according to the programmed weight precision. All plastic weights in our work used 8 bit precision.
Currently, the Loihi's learning rules cannot scalably depend on an extrinsic factor such as modulation.
(Although there exists an external reward modulation variable, these are limited to 4 per core but surrogate gradient learning requires one per neuron.)
In the following paragraph, we describe a workaround for this constraint and the implementation of surrogate gradient learning.

\subsection{Surrogate Gradient Learning in Spiking Neurons}
The main contribution of this work is the implementation of a surrogate gradient-based learning rule in the Loihi research processor using three-factor rules.
Three-factor rules are extensions of Hebbian learning and \ac{STDP}, which can be derived from a normative (top-down) approach \cite{Urbanczik_Senn14_learby}.
Such rules are compatible with a wide number of unsupervised, supervised, and reinforcement learning paradigms \cite{Urbanczik_Senn14_learby}, and implementations can have scaling properties comparable to that of \ac{STDP} \cite{Detorakis_etal18_neursyna}.

The third factor in three-factor rules is motivated by neuromodulators, such as Dopamine, Acetylcholine, or Noradrenaline in reward-based learning \cite{Schultz02_gettform}.
Generally, the weight update of the three-factor learning rule can be written as follows:
\begin{equation}\label{eq:3f_rule}
  \Delta w_{ij}^{3F} \propto f_{pre} f_{post} M_i
\end{equation}
where $f_{pre}$ and $f_{post}$ correspond to functions over presynaptic and post synaptic variables, respectively (for example, but not restricted to traces $P_i$ and $P_j$), and $M_i$ is the modulating term of the postsynaptic neuron $i$.
Note that, to reduce clutter, time indices will be omitted when there is no ambiguity (all terms in the same time step $n$).
The modulating term is a task-dependent function, which can, for example, represent error, surprise, reward, or another variable relevant to the task at hand.

Three-factor rules can be derived from gradient descent on the spiking neural network \cite{Pfister_etal06_optispik,Neftci_etal19_surrgrad}. 
The significance of this compatibility is far-reaching because the techniques and tools of deep learning can be made compatible with training spiking neural networks.
Such rules are often ``local'' in the sense that all the information necessary for computing the gradient is available at the post-synaptic neuron \cite{Neftci18_datapowe}.
For this reason, three-factor rules are a candidate for implementation in neuromorphic hardware.

In our contribution, we start with a three-factor synaptic plasticity rule derived from surrogate gradient descent for classification or regression tasks \cite{Zenke_Ganguli17_supesupe}. 
For the output layer:
\begin{equation}\label{eq:learning_rule}
\begin{split}
\Delta w_{ij} &\propto Error_i~\sigma^\prime(v_i-v_{th}) P_j,\\
\end{split}
\end{equation}
where the $Error_i$ is the instantaneous error 
$(\text{target}_i - \text{output}_i)$, 
$P_j$ is the pre-synaptic trace at the membrane potential, and $\sigma^\prime(v_i[n]-v_{th})$ is the derivative of the smoothed output function with $v_{th}$ the firing threshold.
For the output layer of a classification or regression neural network, this learning rule is equivalent to the Delta rule in artificial neural networks. 
It is derived from the gradient of a \ac{MSE} loss with respect to the neuron parameters, ignoring the refractory variable $R$ and replacing derivatives of $\Theta$ with those of the sigmoid function \cite{Neftci_etal19_surrgrad}.
This rule is the building block of the backpropagation algorithm in multilayer networks, in which case $Error_i$ becomes the backpropagated errors (i.e. the ``deltas'').
Because this rule is derived from gradient descent on the spiking dynamics (i.e. there is no rate-based approximation), it naturally leverages spike timing dynamics.
As a result, this rule and its variations have been shown to perform particularly well on spike-based classification tasks and outperform traditional synaptic plasticity rules such as \ac{STDP} and reward/error modulated \ac{STDP} \cite{Zenke_Ganguli17_supesupe,Kaiser_etal18_synaplas} in equivalent networks.

\subsection{Implementation of Surrogate Gradient Learning in the Loihi Research Processor}\label{sec:loihi3frule}

\begin{figure*}[!t]
\centering
\includegraphics[width=\textwidth]{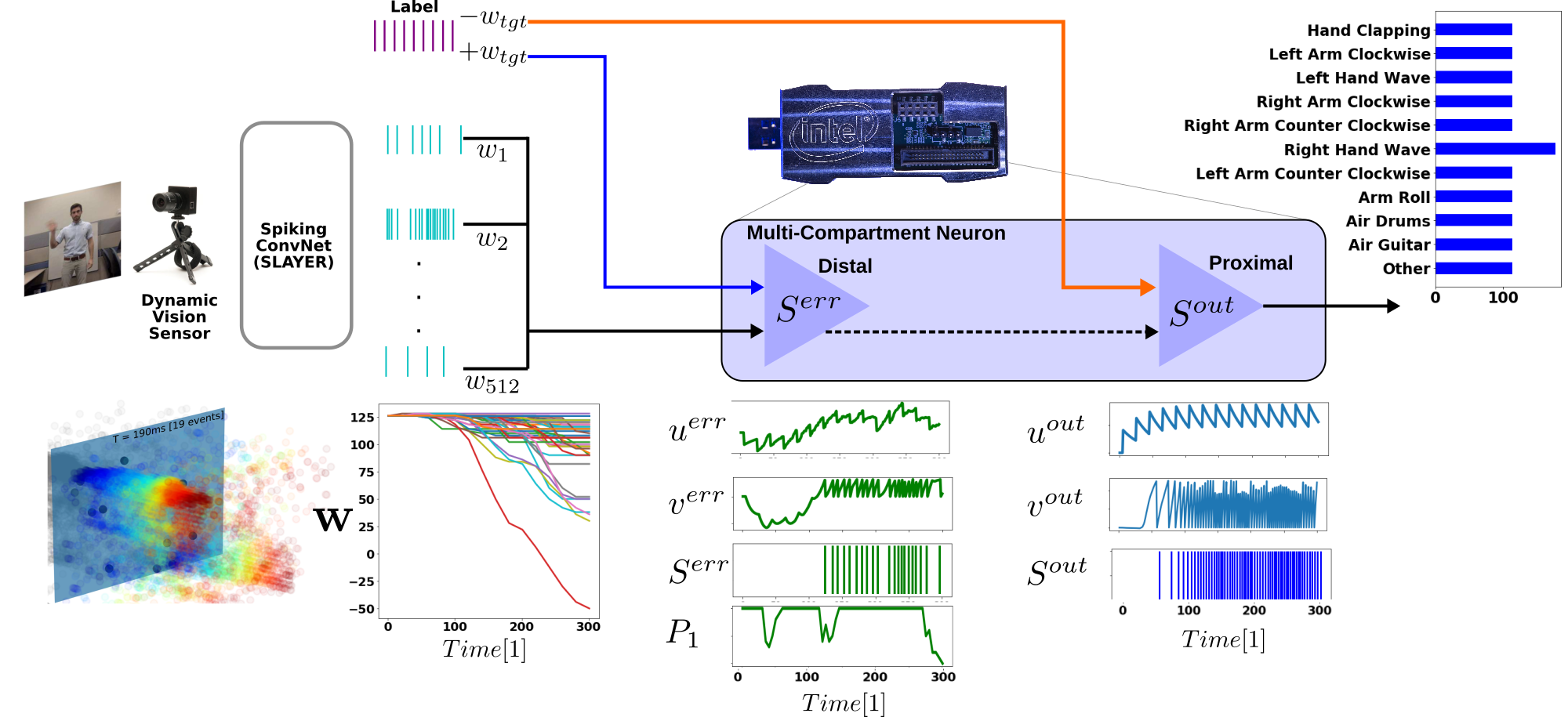}
\caption{
Setup and multi-compartment neuron used in our work. The distal compartment receives spikes from the spiking ConvNet with synaptic weights adjusted by the surrogate gradient rule.
The distal compartment pushes its current to the proximal compartment for integration.
During training, label spikes with weights of opposite signs are received by the distal compartment and proximal compartment.
Error is calculated in the distal compartment as a subtraction of the label input and input from the previous layer.}
\label{fig_sim}
\end{figure*}

Here, we map the Loihi learning dynamics to Eq. (\ref{eq:learning_rule}).
By definition, the factor $\sigma'$ is constrained to the range $[0,1]$. 
It is non-zero when the membrane potential is close to $0$.
Removing this term will cause synaptic weights to endlessly grow, but does not prevent learning in the early stages of training. 
This mode of operation is compatible with the few-shot learning scenario experimented here. 
The two other terms $f_{pre}$ and $M_i$, however, are critical for learning.
Thus, we drop the $\sigma'$ term in (\ref{eq:learning_rule}) and obtain provisionally:
\begin{equation}\label{eq:superspike}
\Delta w_{ij} \propto P_j P^{err}_i.
\end{equation}
where $P^{err}_i$ is an error trace as described below. 
To represent the error, we use dynamics:
\begin{equation}
    \begin{split}
        S_i^{err} &= \Theta(v_i^{err} - v_{th}),\\
        v^{err}_i &= \sum_j w_{ij} P_j - w_{tgt} \hat{P}_i + R^{err}_i + b^{err},
    \end{split}
\end{equation}
where $P$, $\hat{P}$ and $R^{err}_i$ follow the same dynamics as in (\ref{eq:loihi_dynamics}). 
$\hat{P}$ is the input due to the target spikes, and $w_{tgt}>0$ is its weight.
The constant $b^{err}$ is a bias term used to offset the firing rate of the neuron, as explained in the following paragraphs.

The synaptic plasticity processor operates on neural traces computed with pre-synaptic and post-synaptic spike times.
In Loihi, traces are first-order linear filters.
However, as discussed earlier $P$ is a second-order linear filter.
Second-order kernels can be implemented as a subtraction of two first-order kernels \cite{Gerstner_etal14_neurdyna}. 
This subtraction is enabled by the sum-of-products formulation of the plasticity rule (\ref{eq:loihi_rule}).
The error can be implemented with the post-synaptic compartment using the following post-traces:
\[
\begin{split}
    P^{err}_i[n+1] &= \alpha_v P^{err}_i[n] + Q^{err}_i[n],\\
    Q_i^{err}[n+1] &= \alpha_u Q_i^{err}[n] + S^{err}_i[n].
\end{split}
\]
Post-traces can only be positive but errors can be both positive and negative.
This problem can be solved by offsetting the weight updates with a constant term $b$:
\begin{equation}\label{eq:superspike}
\Delta w_{ij} \propto P_j (P^{err}_i - b).
\end{equation}
To enforce the constraint of zero error when output matches target, a constant firing rate is imposed on the error neurons via $b^{err}$. 
In the absence of inputs, this constant firing will create a repeating $P^{err}_i$ trace with average $\langle P^{err}_i \rangle$.
The bias $b$ is then adjusted such that it is equal to the average of this trace:
$
\langle P^{err}_i \rangle = b.
$

Two potential problems remain with this workaround: 
a) bounds on the error magnitude, 
b) Fluctuations in the weight updates even when the error is zero, since only the average update is zero with the baseline firing rate.
Previously work suggest that a) does not prevent convergence in spiking neural network classification tasks \cite{Neftci_etal17_evenranda}.
While (b) is currently unsolved, we observe that it does not prevent few-shot learning.
The learning rule is implemented in Loihi as follows:
\begin{equation}
    \Delta w_{ij} \propto C(Y_1(X_2 - X_1) + X_1 - X_2)
\end{equation}
where $C$ is a constant that captures the learning rate, $X_1$ and $X_2$ are two first-order pre-synaptic traces, $Y_1$ is a first-order post-synaptic trace. 
The $(X_2 - X_1)$ term corresponds to $P_j$ in (\ref{eq:superspike}), and $Y_1$ corresponds to $P^{err}_i$ in (\ref{eq:superspike}). 
Note that to simplify the rule, we approximated $P^{err}$ using a single first-order filter.

\subsection{Multi-compartment Neuron and Readout}
In order to read out the class and to build multilayer learning in future work, we use a multi-compartment neuron.
The neurons consist of two compartments, made of a distal (error) compartment that receives input from a pre-trained network, and a proximal (output) compartment whose voltage integrates the state $u^{err}$ copied from the distal compartment, and does not directly receive spiking input from the convnet: 
\begin{equation}
    \begin{split}
        S_i^{out} &= \Theta(v_i^{out}-v_{th})\\
        P^{out}_i &= \alpha_v P^{out}_i + u^{err}\\
        v^{out}_i &=  P^{out}_i + R^{out}_i + w_{tgt} \hat{P}_i.
    \end{split}
\end{equation}
During training, target spike signals are sent to both the distal compartment and the proximal compartment.
The distal compartment receives positively weighted target spikes to adjust the weight of attached synaptic connections towards a desired class, while the proximal compartment receives equally negative weighted target spikes to cancel out the effect of the label spikes have on the current integrated from the attached distal compartment. 
Targets are not provided during the test phase.
Error encoding and propagation to the proximal compartment takes place in the distal compartment without the need for separate phases. 
Classification accuracy is measured using the spike count in the proximal compartments of the output layer neurons.


\section{Experimental Evaluation}
\subsection{Setup}


The IBM DvsGesture\cite{Amir_etal17_lowpowe} dataset consists of recordings of 29 different individuals performing 10 different actions such as clapping and an unspecified gesture for a total of 11 classes. The actions are recorded  using a DVS camera, an event-based neuromorphic sensor, under three different lighting conditions. The problem is to classify an action sequence video. Samples from the first 23 subjects were used for training and the last 6 subjects were used for testing. The test set contains 264 samples and will be used for the few-shot learning task. Each sample consists of the first 1.45 seconds of the gesture performed. The SNN model trained on the Loihi for DvsGesture classification is described in table \ref{table:arch}. 

\begin{table}[!ht]
\caption{Network architecture}
\label{table:arch}
\centering
\begin{tabular}{|c|c|c||c|} \hline
Layer & Kernel & Output    & Training Method\\ \hline
input & 1a     & 128x128x2 & DVS128 (Sensor)\\\cline{4-4}
1     & 4a     & 32x32x2   & SLAYER (BPTT)\\
2     & 16c5z   & 32x32x16  &\\
3     & 2a     & 16x16x16  &\\
4     & 32c3z   & 16x16x32  &\\
5     & 2a     & 8x8x32    &\\
6     & -      & 512       &\\ \cline{4-4}
out   & -      & 11        & Loihi Plasticity Rule\\ \hline
\end{tabular}
\end{table}

The hidden layers 1 through 6 were trained offline using SLAYER \cite{Shrestha_Orchard18_slayspik} on the Loihi neuron model. 
The output layer synapse weights are trained using the three-factor surrogate gradient plasticity rule described in Sec. \ref{sec:loihi3frule}.

We compare our model to the one used in \cite{Shrestha_Orchard18_slayspik}. 
To assess \cite{Shrestha_Orchard18_slayspik}, the output layer weights are reset and retrained in few-shot learning conditions.
Due to the limited bit precision of the synaptic weights in Loihi (256 levels), the effective learning rate is typically higher than that used in GPUs.
To emulate this, we trained the output layer of the SLAYER model using a higher learning rate than the original for a better comparison.  
In our few-shot learning experiments, using a higher learning rate achieves a higher accuracy because the training data is small and presented only once. 
We also compare our results to a model trained using SLAYER with hidden layers pre-trained SLAYER network but with the output layer replaced with a fully connected linear layer trained using the \ac{MSE} of the membrane potentials of the neurons instead of classification with spike counts. 
\par Because the hidden layers of each model are pre-trained on the DvsGesture training data, 
we train and test the models on the DvsGesture test dataset. 
Using the test dataset we set up three N-way classification tasks with 1, 5, and 14 shots.
N-way classification involves selecting N unseen classes, provide the model with K different instances, also known as shots, of each of the N classes, and evaluate the model's ability to classify new instances within the N classes.
The test data is split into separate training and testing sets.
The training set consists of N$\cdot$K samples taken from the first 154 samples of the DvsGesture Test dataset.
During training, we present the data of the K shots for a single learning epoch.
The N-way classification models are then tested on the last 110 samples of the DvsGesture Test dataset not used during training. 
\par Additionally, we set up three M+N-way classification tasks with 1, 5, and 20 shots.
For M+N-way classification, we pre-trained the SLAYER network offline using only M classes of the DvsGesture Training dataset.
We then reset the output layer weights and retrain them using the surrogate gradient plasticity rule on Loihi on N unseen classes, provide the model with K shots of each of the N classes, and evaluate the model's ability to classify new instances within the N classes.
The M+N-way classification models are trained using the DvsGesture Training dataset, and tested using the DvsGesture Test dataset.

\subsection{Results}
We present the $N$-way classification results in Table \ref{table:dvs_table}, and the $M+N$-way classification results in Table \ref{table:dvs_table5-way}. 
Each model is trained with K shots presented a single time, and then tested on samples of each class held out from training.
We use $N=11$ for the number of classes for the $N$-way classification experiments, which is the number of classes in DvsGesture.
For the $M+N$-way classification experiment, we use $M=6$ for the number of classes pre-trained offline using SLAYER, and $N=5$ for the number of classes trained on the network implemented on Loihi using the surrogate gradient plasticity rule.
``Train'' refers to classification accuracy on training samples using saved weights with plasticity disabled.
``Test'' is the classification accuracy on the samples held out of the training procedure.
For each type of model, the same training and testing data samples were used.
The SLAYER+Linear section of Table \ref{table:dvs_table} shows the results of the model consisting of a pre-trained SLAYER hidden layer connected to an ANN Linear output layer that uses \ac{MSE} loss, measuring the error of target versus output membrane potential quantities rather than a classification accuracy using the number of spikes over the presentation time. 

The surrogate gradient plasticity rule implemented on Loihi outperforms the purely SLAYER model on few-shot learning tasks in both the 11-way classification experiments, and the 6+5-way classification experiments.
In the 11-way few-shot classification experiments, the plasticity rule achieved $32.1\%$ better accuracy at test time than the SLAYER rule of the SLAYER+Spiking model for one-shot learning, with the plasticity rule achieving better accuracy on the 5 and 14 shot learning tasks as well.
Interestingly, the SLAYER+Linear model achieved $21.7\%$ better test accuracy than the SLAYER+Spiking model on one-shot learning, but performed poorer on the 5 and 14 shot learning tasks than the SLAYER+Spiking model.
These results indicate that the plasticity rule augmented with the pre-trained features does not need large amounts of training data to generalize. 
The performance of the plasticity rule in the $6+5$-way experiments was not as high as the 11-way experiments because the 5 unseen classes were unseen by the entire network and not just unseen by the output layer, but did outperform the $6+5$-way SLAYER models.
The plasticity rule achieved $24.5\%$ better accuracy at test time than the SLAYER rule of the SLAYER+Spiking model and $23.2\%$ better accuracy than the SLAYER+Linear model for 20 shot learning, with the plasticity rule achieving better accuracy on the 1 and 5 shot learning tasks as well.
These results indicate that the plasticity rule augmented with features pre-trained on some classes but not all can generalize to completely new classes of data.

\begin{table}[!t]
\centering
\caption{\label{table:dvs_table}11-way Few-shot classification on the DvsGesture dataset}
\begin{tabular}{|c|c|c|c|c|}
\hline
Dataset                         & Learning Method                                   & Shots       & Train           & Test            \\ \hline
\multirow{9}{*}{DVSGesture    } & \multirow{3}{*}{Loihi Plasticity Rule}            & \textbf{1}  & \textbf{100\%}  & \textbf{52.2\%} \\ \cline{3-5}
                                &                                                   & \textbf{5}  & \textbf{86.6\%} & \textbf{56.8\%} \\ \cline{3-5}
                                &                                                   & \textbf{14} & \textbf{72.2\%} & \textbf{64.7\%} \\ \cline{2-5}
                                & \multirow{3}{*}{SLAYER+Spiking}                   & 1           & 9.1\%           & 20.1\%          \\ \cline{3-5}
                                &                                                   & 5           & 40\%            & 50.1\%          \\ \cline{3-5}
                                &                                                   & 14          & 56.5\%          & 61.8\%          \\ \cline{2-5}
                                & \multirow{3}{*}{SLAYER+Linear }                   & 1           & $<$1\%             & 41.8\%          \\ \cline{3-5}
                                &                                                   & 5           & 38.2\%          & 42.7\%          \\ \cline{3-5}
                                &                                                   & 14          & 53.9\%          & 51.8\%          \\ \hline
\end{tabular}
\end{table}


\begin{table}[!t]
\centering
\caption{\label{table:dvs_table5-way}6+5-way few-shot classification on the DvsGesture dataset 
}
\begin{tabular}{|c|c|c|c|c|}
\hline
Dataset                         & Learning Method                                   & Shots       & Train           & Test            \\ \hline
\multirow{9}{*}{DVSGesture    } & \multirow{3}{*}{Loihi Plasticity Rule}            & \textbf{1}  & \textbf{40\%}   & \textbf{40\%}   \\ \cline{3-5}
                                &                                                   & \textbf{5}  & \textbf{60\%}   & \textbf{43.3\%} \\ \cline{3-5}
                                &                                                   & \textbf{20} & \textbf{73.5\%} & \textbf{56.2\%} \\ \cline{2-5}
                                & \multirow{3}{*}{SLAYER+Spiking}                   & 1           & 40\%            & 35\%            \\ \cline{3-5}
                                &                                                   & 5           & 20\%            & 41.7\%          \\ \cline{3-5}
                                &                                                   & 20          & 37\%            & 31.7\%          \\ \cline{2-5}
                                & \multirow{3}{*}{SLAYER+Linear }                   & 1           & 40\%            & 33\%            \\ \cline{3-5}
                                &                                                   & 5           & 36\%            & 27.5\%          \\ \cline{3-5}
                                &                                                   & 20          & 35\%            & 33\%            \\ \hline
\end{tabular}
\end{table}


\section{Conclusion}
 The results achieved in the few-shot learning experiment demonstrate the building blocks for online multi-layer learning and transfer learning in neuromorphic hardware.
Neuromorphic hardware that can achieve high accuracy on only a few observations is a long-sought feature in applications where large datasets are nonexistent, cannot be accessed or where energy is limited, such as in robotic and embedded IoT systems.
However, there are still challenges for achieving highly accurate continual ``life-long'' learning in neuromorphic hardware such as catastrophic forgetting \cite{McClelland_etal95_whyther}.
Future work will address these challenges by exploring synaptic consolidation dynamics \cite{Kirkpatrick_etal17_overcata,Benna_Fusi15_compprin} and reducing the variance of the learning when no error is made (problem (b) in the text).
The solution described here trains the final layer only and thus foregoes the deep credit assignment problem. 
Our ongoing and future work is addressing the deep credit assignment using the technique of local losses \cite{Kaiser_etal18_synaplas}, thus allowing to train multiple layers simultaneously. 
As the plasticity rule is improved, the applicability is likely to extend beyond classification, namely unsupervised learning and reinforcement learning tasks.

\section{Acknowledgments}
This work was supported by Intel Corporation (KMS); the National Science Foundation under grant 1652159 (EON); the National Science Foundation under grant 1640081 (EON), and the Nanoelectronics Research Corporation (NERC), a wholly-owned subsidiary of the Semiconductor Research Corporation (SRC), through Extremely Energy Efficient Collective Electronics (EXCEL), an SRC-NRI Nanoelectronics Research Initiative under Research Task ID 2698.003 (EON, KMS).




%
\bibliographystyle{IEEEtran}
\bibliography{biblio_unique_alt,extra_bib}

\end{document}